\documentclass[a4paper,twoside]{article}

\usepackage{epsfig}
\usepackage{subcaption}
\usepackage{calc}
\usepackage{amssymb}
\usepackage{amstext}
\usepackage{amsmath}
\usepackage{amsthm}
\usepackage{multicol}
\usepackage{pslatex}
\usepackage{apalike}
\usepackage{algorithm2e}
\usepackage[bottom]{footmisc}
\usepackage{SCITEPRESS}     
\usepackage{graphicx}

\begin{document}

\title{CSE: Surface Anomaly Detection with Contrastively Selected Embedding}

\author{\authorname{Simon Thomine\sup{1,2}, Hichem Snoussi\sup{1}}
\affiliation{\sup{1}\textbf{University of technology Troyes, Troyes, France }}
\affiliation{\sup{2}\textbf{AQUILAE, Troyes, France}}
\email{simon.thomine@utt.fr}
}


\abstract{
Detecting surface anomalies of industrial materials poses a significant challenge within a myriad of industrial manufacturing processes.
In recent times, various methodologies have emerged, capitalizing on the advantages of employing a network pre-trained on natural images for the extraction of representative features. Subsequently, these features are subjected to processing through a diverse range of techniques including memory banks, normalizing flow, and knowledge distillation, which have exhibited exceptional accuracy.
This paper revisits approaches based on pre-trained features by introducing a novel method centered on target-specific embedding. To capture the most representative features of the texture under consideration, we employ a variant of a contrastive training procedure that incorporates both artificially generated defective samples and anomaly-free samples during training.
Exploiting the intrinsic properties of surfaces, we derived a meaningful representation from the defect-free samples during training, facilitating a straightforward yet effective calculation of anomaly scores.
The experiments conducted on the MVTEC AD and TILDA datasets demonstrate the competitiveness of our approach compared to state-of-the-art methods.
}

\maketitle

\section{\uppercase{Introduction}}
\label{sec:introduction}

The unsupervised anomaly detection domain, especially in industrial applications, has attracted considerable attention in the past few years. Convolutional Neural Networks (CNNs) have emerged as a significant breakthrough in this field by introducing effective mechanisms for anomaly detection.
The efficacy of CNNs resides in their capacity to analyze and process visual data, including images and surfaces, through the capture of spatial features and patterns.
Deep learning has gained increasing momentum in the industry owing to its capacity to derive intricate representations from extensive datasets, adapt to diverse domains, and execute real-time processing.
Harnessing the potential of deep learning enables industries to attain heightened accuracy, automation, and efficiency across diverse applications, including the detection of anomalies in quality control.

\noindent
In the industrial setting, where precision and accuracy are of paramount importance, it is imperative to employ specialized and faultless methods that adhere to stringent standards, minimizing errors and ensuring flawless performance tailored to the specific requirements of the environment.

\noindent Recently, there has been a proliferation of approaches capitalizing on extracted features derived from pre-trained classifiers. These classifiers, trained on extensive databases like ImageNet \cite{krizhevsky_imagenet_2017}, encapsulate a wealth of informative features at various levels, encompassing both low-level details such as contours and color, as well as higher-level features that are more contextual and abstract in nature.

\noindent These approaches regroups mainly memory banks, normalizing flows and knowledge distillation that all offers impressive results while guaranteeing a decent inference time. The purpose of this paper is to introduce a new method based on pre-trained features that broadens the possibilities in terms of approaches to handle this specific problem while concurrently minimizing inference time.

\begin{figure*}[h]
\captionsetup{font=small,justification=centering}
\centerline{\includegraphics[scale=0.26]{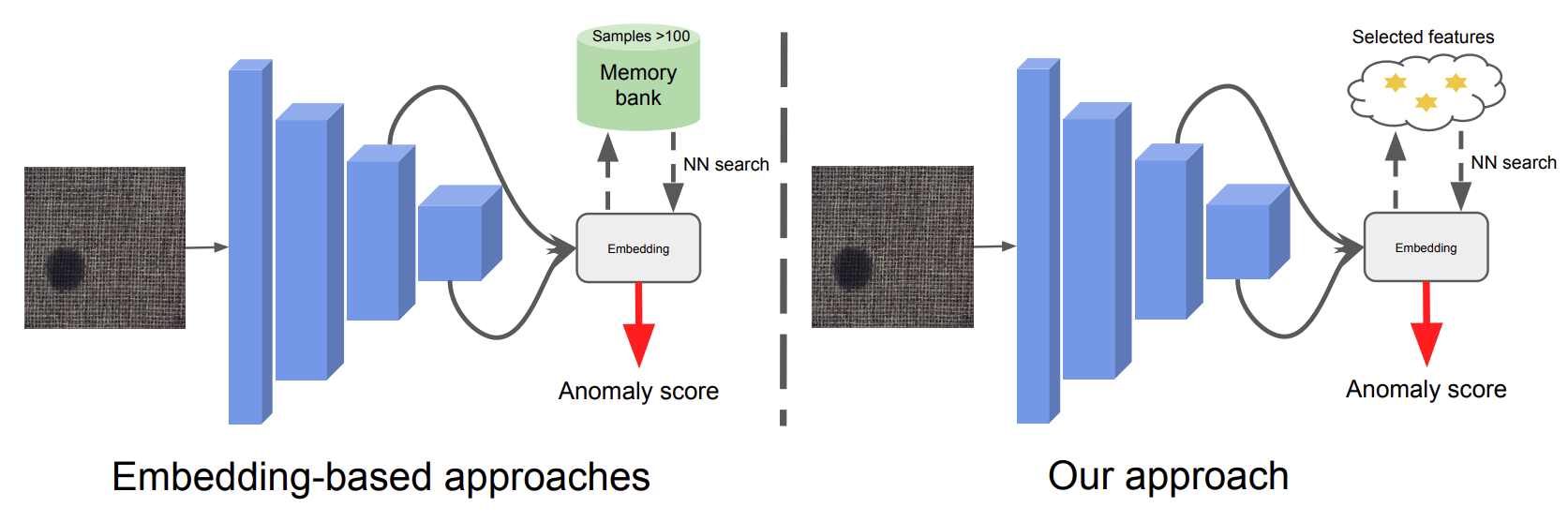}}
\renewcommand{\arraystretch}{1}
\caption{ 
A comprehensive examination of the distinctions between our methodology and alternative embedding-based approaches during the inference phase. Limiting the comparison to a few specifically chosen samples, instead of encompassing the entire set of features, results in a considerable reduction in inference time.}
\label{Fig.1}
\end{figure*}

\noindent The primary objective of feature extraction from pre-trained models is to compile the most representative features of the object, emphasizing those that exhibit differences in the presence of an anomaly. Conventional approaches employ various strategies for feature extraction, including sub-sampling of features, normalizing flows, or reconstruction-based approaches. 
Our conviction lies in the idea that, for effective anomaly detection, guiding the model toward features with optimal "anomaly detection" capabilities for our target texture is crucial.
To this end, we employ a defect generation method, such as the one introduced in DRAEM \cite{zavrtanik_draem_2021}, to assist the model in extracting features that are responsive to defects.
\noindent Our model comprises three primary components: a pre-trained feature extractor, an embedder/encoder responsible for aggregating the most representative features, and a decoder designed to avoid a trivial embedded representation. 
In the process of training the model, two samples are subjected to processing: one being anomaly-free, and the other exhibiting either an absence of anomalies or the presence of an artificially generated defect with a specified probability. Subsequently, the cosine similarity measure is employed as a contrastive loss function, with the objective of minimizing the embedding distance between the two samples if both are anomaly-free, or increasing it otherwise. The anomaly-free embedding of the defect-free sample is then subjected to the decoder to minimize the reconstruction loss, thereby enhancing the diversity of the embedding representation. 
Following the completion of the training process, a k-means clustering procedure is implemented to extract a predetermined number of clusters, which subsequently functions as a feature bank. In the testing phase, the anomaly score is computed efficiently and accurately by comparing these clusters with the embedding of the test sample.
Figure \ref{Fig.1} described our proposed score calculation approach compared to other embedding-based approaches.

\noindent The primary contributions of this paper are outlined as follows:
\begin{itemize}
  \item  An embedder capturing the most representative features of a target surface through the application of a contrastive training approach, showcasing exceptional performance in the domain of texture defect detection and achieving state-of-the-art capabilities.
  \item A contrastive cosine loss formulated with the intention of amplifying the difference in embedding representation between defective samples and anomaly-free samples, while simultaneously diminishing this difference between two anomaly-free samples.
  \item  A comprehensive training design incorporating a decoder to augment the variability of the embedded features, thereby preventing a trivial representation.
  \item A k-means clustering approach extracting the most significant clusters for anomaly scoring.
\end{itemize}

\noindent Subsequent to the introductory section, the following segment of this manuscript is devoted to a comprehensive review of existing literature concerning deep learning methodologies utilized in unsupervised industrial anomaly detection. Section 3 presents our innovative approach with a precise description of each components. Section 4 is dedicated to a series of experiments to evaluate the efficacy of our proposed model. In section 5, an ablation study is conducted to present the benefits of each components from the contrastive approach relevance to a comparison between training methods for the decoder along with an explanation of the choice of features. A conclusive section offers a summary of the paper's findings, outlines the limitations and proposes potential avenues for future research.

\section{\uppercase{Related Work}}

In the realm of industrial applications, the comprehensive compilation of data pertaining to every potential defect in an object or texture poses a challenging and time-intensive task where neglecting to account for all types of defects can result in sub-optimal performance outcomes \cite{han_adbench_2022}.
This section provides a thorough overview of methodologies for unsupervised anomaly detection, placing specific emphasis on recent advancements that leverage deep learning techniques.

\noindent In early literature, generative models like auto-encoders \cite{mei_automatic_2018,nguyen_gee_2019,zavrtanik_draem_2021}, generative adversarial networks \cite{goodfellow_generative_2014} , and their variations \cite{schlegl_f-anogan_2019,pourreza_g2d_2021,liang_omni-frequency_2022} were employed to reconstruct normal images from anomalous ones.
Notwithstanding their utility, these methods encountered difficulties in accurately reconstructing complex objects or surfaces, occasionally leading to the generation of faulty samples.

\noindent In recent times, there has been a growing conviction that exploiting fine-grained visual features can contribute significantly to advancements in anomaly detection. Responding to this conjecture, emerging methodologies prioritize the extraction of representations from normal samples, and a prevailing approach in anomaly detection involves utilizing models pre-trained of external images datasets to comprehend the distribution of normal features.

\noindent The utilization of features extracted from pre-trained networks, especially those trained on extensive datasets such as ImageNet \cite{deng_imagenet_2009}, has been observed to confer superior anomaly detection accuracy when compared to the direct processing of the image itself.

\noindent Within this framework, three predominant methods have emerged to exploit the extracted features.

\noindent One method focuses on estimating the distribution of the normal pattern within a parametric framework, particularly by employing normalizing flows \cite{rezende_variational_2016}. In the training phase, flow-based models aim to minimize the negative log-likelihood loss associated with normal images, aligning their features with the target distribution to enhance the performance of the anomaly detection system. Various strategies were employed to improve performance, including the utilization of a 2D flow \cite{yu_fastflow_2021} or the adoption of a cross-scale flow \cite{rudolph_fully_2021}.

\noindent Alternative approaches employed the concept of knowledge distillation \cite{hinton_distilling_2015} adapted to unsupervised anomaly detection. In this approach, a student network is trained on normal samples, employing the output features of a pre-trained teacher network initially designed for classification tasks. In the testing phase, the objective of the student network is to emulate the output features of the teacher network when given defect-free samples. Nevertheless, its accuracy declines when confronted with defective samples, facilitating the derivation of a meaningful anomaly score. Diverse methods have emerged based on this paradigm such as a multi-layer feature selection \cite{wang_student-teacher_2021}, a reverse distillation approach \cite{deng_anomaly_2022} \cite{tien_revisiting_nodate} or a mixed-teacher approach \cite{thomine_mixedteacher_2023}.

\noindent Memory banks approaches rely on diverse defect-free samples to accumulate pertinent features, thereby establishing a bank of features dedicated to the comparison with new samples.
PatchCore \cite{roth_towards_2021} uses a pre-trained classifier to extract specific layers and then gathers features based on their awareness and sub-samples these features. Subsequently, these features are deposited in a memory bank, and the detection of anomalies is accomplished by comparing patch-level distances between the core set and a given sample. Nonetheless, it is crucial to acknowledge that these methods face limitations when trained on extensive datasets, as they demand significant computational resources for the establishment of memory banks and necessitate intricate architectural considerations.

\noindent Other approaches rely on the generation of custom defects. Significantly, the DRAEM method \cite{zavrtanik_draem_2021}, introduces a discriminatively trained autoencoder to generate textural defects using the DTD (Describable Textures Dataset) dataset \cite{cimpoi_describing_2014} and Perlin noise. The CutPaste \cite{li_cutpaste_2021} and MemSeg \cite{yang_memseg_nodate} approaches have also suggested the generation of structural defects to introduce diversity into the defect pool. The employed methodologies demonstrate exceptional outcomes and hold promise for textural anomaly detection, given the inherent properties of surfaces that render the generation of defects comparatively more straightforward.

\section{\uppercase{Proposed Method}}

\begin{figure*}[h]
\captionsetup{font=small,justification=centering}
\centerline{\includegraphics[scale=0.24]{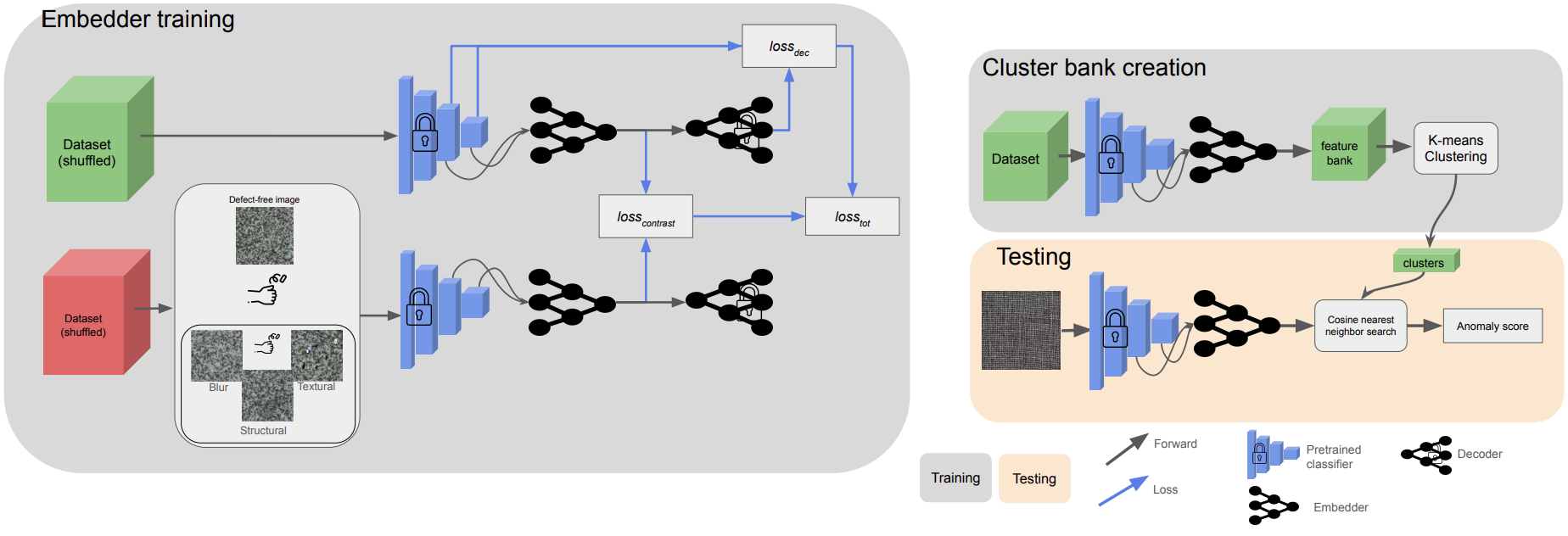}}
\renewcommand{\arraystretch}{1}
\caption{ 
The complete training process. The training of the embedder constitutes the initial step, followed by the computation of clusters derived from the embedding representations. }
\label{Fig.3}
\end{figure*}

This section is devoted to delineating our proposed methodology, which capitalizes on distinct subcomponents to achieve efficient training and precise outcomes. Our approach relies on a contrastive training process that exploits synthesized anomalies and utilizes deep features extracted from a pre-trained model to derive a precise embedding. The complete architecture is shown in Figure \ref{Fig.3}.

\subsection{Image corruption with synthesized anomalies}
\begin{figure}[h]
\captionsetup{font=small,justification=centering}
\centerline{\includegraphics[scale=0.12]{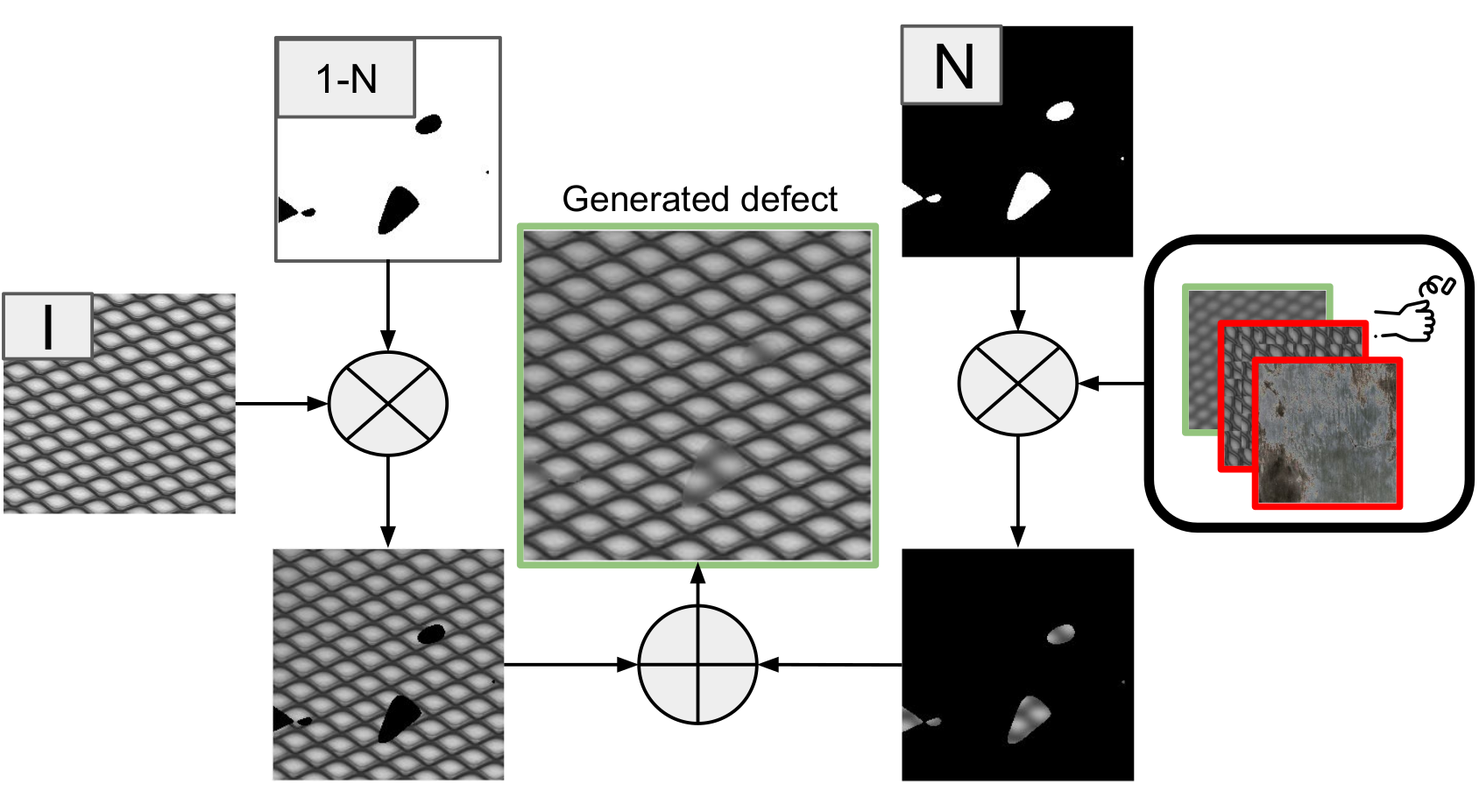}}
\renewcommand{\arraystretch}{1}
\caption{ 
The defect generation process. N is the mask generated by thresholding a Perlin noise and (1-N) denote its negation. I is the original image.}
\label{Fig.3.1}
\end{figure}

To conduct contrastive training, it is imperative to generate anomalies. In alignment with contemporary literature, our anomaly detection process is based on Perlin Noise generation and encompasses various types of anomalies, including structural anomalies \cite{yang_memseg_nodate}, textural anomalies utilizing the DTD dataset \cite{zavrtanik_draem_2021} \cite{cimpoi_describing_2014}, and a novel blurry noise introduced through a straightforward application of Gaussian noise with a randomly generated kernel applied to the original image. 
The complete process of defect generation is detailed in Figure \ref{Fig.3.1}.
Every category of defect manifests with equal probability during the training process to ensure a balanced training regimen and prevent bias towards any particular anomaly type.
It is crucial to note that defects are randomly generated during the training process rather than pre-existing before training. This approach aims to mitigate overfitting and enhance the model's capacity to effectively address a diverse range of defects. 

\subsection{Anomaly detection specific embedding}

To achieve efficient defect detection, the embedding is trained through a contrastive process, wherein the embedder is presented with pairs of images. These pairs consist of either two defect-free samples or one anomalous sample paired with one defect-free sample. Each scenario occurs with equal probability. Subsequently, the embedder is trained to augment the dissimilarity between features for antagonistic samples, while reducing it for correct samples.
 
\noindent In the context of surfaces, conducting contrastive training poses challenges, as a texture with a minor defect remains highly similar to a defect-free texture. To alleviate this issue, we opted to train our feature embedder using deep features extracted from a pre-trained model. Deep features offer the advantage of possessing a substantial receptive field and a relatively low resolution. Consequently, the features of a defective sample are highly likely to encompass a substantial portion of the image.

\noindent To retain spatial information and simplify the embedder architecture, we opted to exclusively employ convolutions with a kernel size of one. For enhanced capabilities, the embedder possesses the capacity to utilize features from various deep layers and efficiently fuse them without incurring any additional inference time cost.

\noindent Given a training dataset of images without anomaly ${D=[I_1,I_2,...,I_n]}$, our goal is to extract the relevant feature from the $L$ top layers of a pre-trained model. For an image ${I_k} \in R^{w\times h\times c}$ where $w$ is the width, $h$ the height, $c$ the number of channels and $l$ the $l^{th}$ bottom layer, the output features are noted as $F^l(I_k) \in R^{w_l\times h_l\times c_l}$. 
The embedded feature is denoted as $E(I_k)$, signifying the embedding of the features extracted from the image $I_k$ by the pre-trained model.
When presented with another image $I_m$, our aim is to enhance the disparity between $E(I_k)$ and $E(I_m)$ in the case of a defective $I_m$, while reducing this difference if $I_m$ is non-defective.

\noindent The design of the embedder is straightforward, featuring a sequence of pointwise convolution layers, complemented by a ReLU layer, a batch normalization layer, and culminating in an average pooling layer that acts as a smoothing component. In the event of input features from multiple layers, the features are initially upscaled to match the size of the largest features and subsequently concatenated before being fed into the embedder.

\subsection{Contrastive cosine loss}
Our contrastive loss relies on cosine similarity, as opposed to the conventional mean square error. This choice is driven by the superior results observed and the absence of a margin parameter, which can be challenging to optimize.
The cosine similarity is defined as:  
\begin{equation}
CosSim(E(I_k),E(I_m))=\frac{E(I_k) \cdot E(I_m)}{\lVert E(I_k)) \rVert \lVert E(I_m)\rVert}
\label{eq.5}
\end{equation}

The cosine contrastive loss function is defined as: \\

\begin{equation}
    \text{loss}_{\text{contr}} =
    \scalebox{0.75}{
        $\begin{cases}
            1 + CosSim(E(I_k),E(I_m)) & \text{if $I_m$ is defective}\\
            1 - CosSim(E(I_k),E(I_m)) & \text{otherwise} 
        \end{cases}$
    }
\end{equation}

where $CosineSim(E(I_k),E(I_m)) \in [-1;1]$.
The objective of this loss function is to enhance the similarity of features from defect-free samples and amplify the discrepancy between features otherwise.

\subsection{Decoder loss}

During the training of our model using only the contrastive loss, we encountered an issue of trivial representation in our embedding. This manifested as all embedded features being identical to each other. This phenomenon is attributed to the absence of diversity requirements in the training objective. 
To mitigate this phenomenon, we introduced a decoder designed to reconstruct features from the embedder dimension to the original dimension. The objective was to ensure diversity, as the decoder would be unable to reconstruct the original dimension from a trivial representation.
Significant to note is that the decoder remains untrained throughout the training process and is initialized with random weights. Further details on this aspect are elaborated in the ablation study.
This decoder process is done only on the defect-free image $I_k$ and the reconstruction of the layer $l$ is noted as $R^l(I_k)$. \\

\begin{figure*}[h]
\captionsetup{font=small,justification=centering}
\centerline{\includegraphics[scale=0.19]{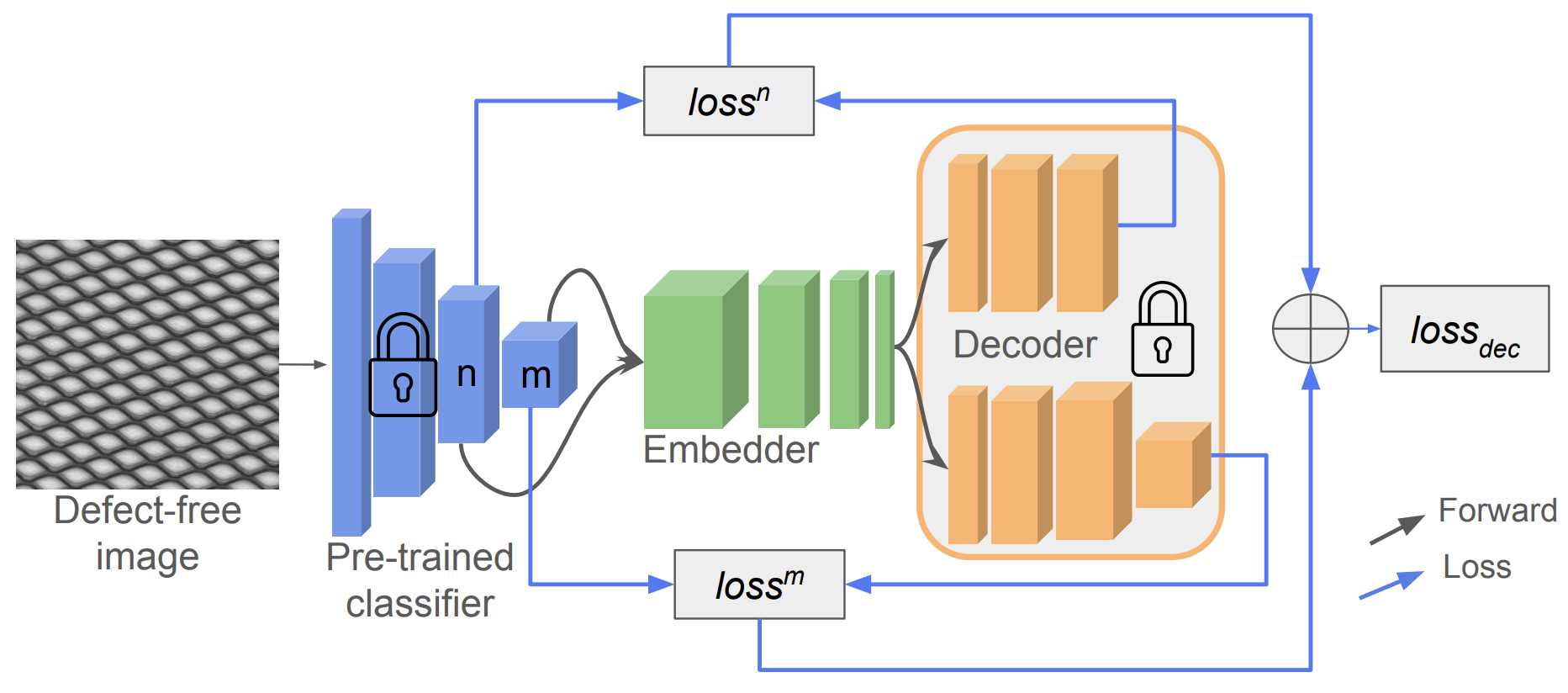}}
\renewcommand{\arraystretch}{1}
\caption{ 
The decoder process for multi-layer embedder. Throughout the training process, both the pre-trained classifier and the decoder remain in a frozen state.}
\label{Fig.3.2}
\end{figure*}

The pixel-loss function is defined as : \\
\begin{equation}
ploss^{l}(I_k)_{ij}=\frac{1}{2}\lVert F^l(I_k)_{ij}-R^l(I_k)_{ij} \rVert 
\label{eq.1}
\end{equation}
with $ploss^{l} \in \mathbb{R}^{H_l\times W_l}$,the layer l loss function as :
\begin{equation}
loss^{l}(I_k)=\frac{1}{w_lh_l}  \sum_{i=1}^{w_l} \sum_{j=1}^{h_l} ploss^{l}(I_k)_{ij}  
\label{eq.2}
\end{equation}
\noindent  and the decoder loss is written as: 
\begin{equation}
loss_{dec}(I_k)= \sum^{l} loss^{l}(I_k) 
\label{eq.3}
\end{equation}
The decoder process is described in Figure \ref{Fig.3.2}.

\noindent Ultimately, the total loss can be expressed as:
\begin{equation}
loss_{tot}(I_k)= loss_{dec}(I_k) + \alpha \cdot loss_{contr}(I_k) 
\label{eq.4}
\end{equation}
with $\alpha$ the weighting factor. In our experimental setup, $\alpha$ is configured to 10.

\noindent A description of the decoder architecture for multiple layers can be seen in Figure \ref{Fig.3.2}.

\subsection{Anomaly scoring and memory bank}
Cutting-edge memory bank methodologies necessitate the utilization of a memory bank whose scale aligns with that of the training dataset, thereby maximizing accuracy.
By depending on shallow and mid-level features, these methodologies necessitate a larger number of defect-free samples to enhance the likelihood of aligning with the features of a defect-free sample during the inference process. In contrast, leveraging deep features and concentrating on surfaces obviates the requirement for a comprehensive memory bank, as features characterized by a high level of abstraction lack fine-grained details such as edges and contours. 
To obtain computable features for deriving an anomaly score, we employed the k-means algorithm on the embeddings of all elements within the defect-free training dataset, utilizing a variable number of clusters based on the texture's diversity. In pursuit of a domain-generalized approach, a greater number of clusters may be employed compared to a texture characterized by regular samples.
In our experiments with public datasets, we configured the number of clusters to one, thereby rendering our cluster equivalent to the computation of the mean of defect-free training samples.
The anomaly score is subsequently determined by calculating the cosine similarity with all clusters and selecting the minimum distance.
The process is described in Figure \ref{Fig.3}.

\section{\uppercase{Experiments}}
\label{sec:conclusion}

\subsection{Implementation details}
We used the deep layers of an EfficientNet-b3 \cite{tan_efficientnet_2020} pre-trained on ImageNet as pre-trained extractor. The training and inference processes were conducted on an RTX 3090ti. 
In order to maintain consistency with other unsupervised approaches during the evaluation process, either the images were resized to 256x256 pixels and then further processed through center-cropping to a final size of 224x224 pixels for the dataset MVTEC AD, or conducted the evaluation under identical conditions using the anomalib library \cite{akcay_anomalib_2022} for the TILDA \cite{deutsche_forschungsgemeinschaft_tilda_nodate} dataset.
During training, the dataset was split into a training set, comprising 70\% of the data, and a validation set, containing the remaining 30\%. Throughout the training phase, we systematically tracked the validation loss, preserving the checkpoint corresponding to the minimum recorded loss value.
To optimize the model's parameters, we utilized the ADAM optimizer \cite{kingma_adam_2017} with a learning rate of 0.0004. To expedite convergence, we implemented a one-cycle learning rate scheduler \cite{smith_disciplined_2018} and conducted training over 100 epochs, utilizing a batch size of 8.

\noindent All experiments presented were conducted utilizing the deep layers of EfficientNet-B3, employing input sizes of 136x14x14 and 384x7x7, along with an embedding dimension set at 64x7x7.

\subsection{Experiments on surface datasets}
We used the area under the receiver operating characteristic curve (AUROC) to assess the image-level anomaly detection performance.

\noindent Our evaluation was conducted in different surfaces datasets namely the MVTEC AD dataset \cite{bergmann_mvtec_2019} and the TILDA dataset \cite{xie_improved_2021}. These datasets compile a substantial amount of textural samples representing various conceivable scenarios.  

\subsubsection{MVTEC AD surfaces}
\begin{figure}[h]
\captionsetup{font=small,justification=centering}
\centerline{\includegraphics[scale=0.14]{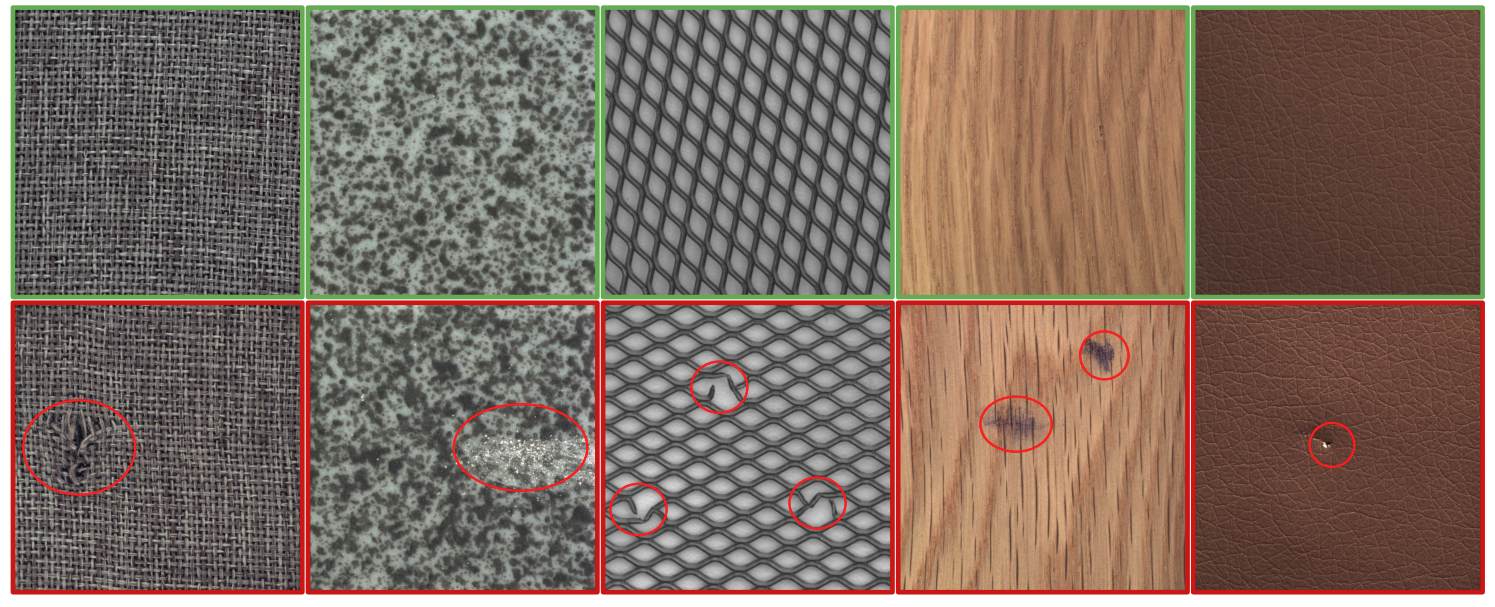}}
\renewcommand{\arraystretch}{1}
\caption{ 
An overview of MVTEC AD surfaces. The figure's upper section contains defect-free samples, whereas defective samples are situated in the lower part. Red encirclement highlights the defects.}
\label{Fig.Mvtec}
\end{figure}

The widely recognized and demanding benchmark MVTEC dataset gathers 5 surfaces and 10 objects in the realm of industrial inspection. Since our method is designed for unsupervised surface defect detection, we evaluate only on the 5 surfaces. An overview of the dataset is shown in \ref{Fig.Mvtec}. The results of our evaluation are depicted in Table \ref{Table 1}.

\begin{table*}[!h]
\captionsetup{font=small,justification=centering}
\caption{Anomaly detection results with AUROC on MVTEC surfaces}
\resizebox{15.5cm}{!}{
\begin{tabular}{|c|ccccc|c|} 
 \hline
		 \textbf{Category} & \textbf{CFA \cite{lee_cfa_2022}} & \textbf{PatchCore \cite{roth_towards_2021}}& \textbf{FastFlow \cite{yu_fastflow_2021}}&
        \textbf{RD++ \cite{tien_revisiting_nodate}} & \textbf{MixedTeacher \cite{thomine_mixedteacher_2023}} & \textbf{Ours} \\
        \hline
		{carpet} &
		97.3  & 98.7 & 99.4 & \textbf{100} & 99.8 & \textbf{100} \\
			
    	{tile} &
		99.4  & 98.7 & \textbf{100} & 99.7 & \textbf{100} & 99.3 \\
		
		{wood} &
		\textbf{99.7}  & 99.2 & 99.2 & 99.3 & 99.6 & \textbf{100} \\
		
		{leather} &
		  \textbf{100}  & \textbf{100} & 99.9 & \textbf{100} & \textbf{100} & \textbf{100} \\

        {grid} &
		  99.2  & 98.2 & \textbf{100} & \textbf{100} & 99.7 & 99.6 \\
		
		\hline
		{Mean} &
		99.1  & 99.0 & 99.7 & \textbf{99.8} & \textbf{99.8} & \textbf{99.8} \\
		
	    \hline
\end{tabular}
}

\label{Table 1}
\end{table*}

\noindent Table \ref{Table 1} illustrates the competitive efficacy of our methodology relative to contemporary approaches, exhibiting a mean Area Under the Receiver Operating Characteristic (AUROC) comparable to leading models and demonstrating state-of-the-art performance on wood surface.

\subsubsection{TILDA dataset}
\begin{figure}[h]
\captionsetup{font=small,justification=centering}
\centerline{\includegraphics[scale=0.12]{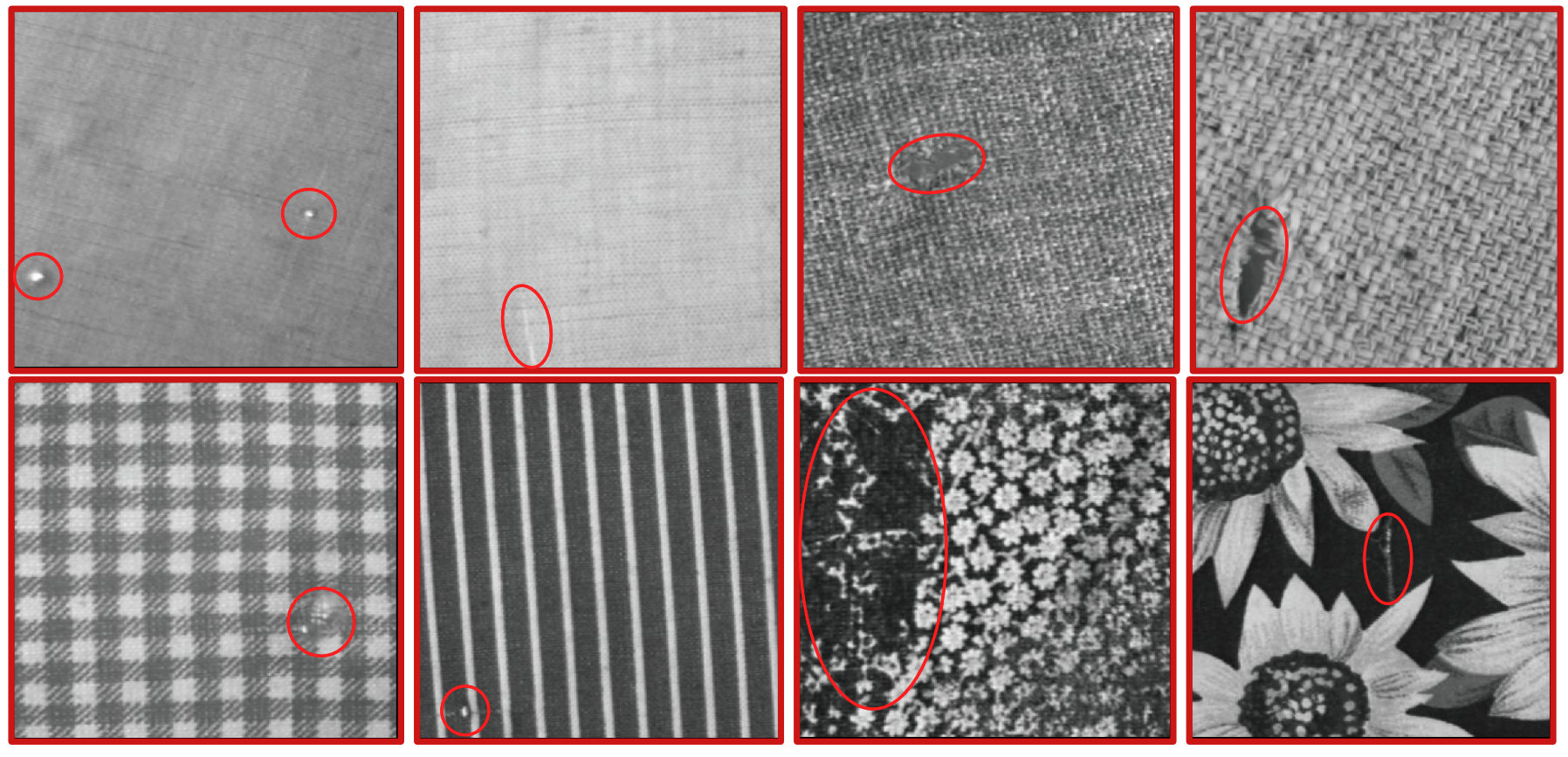}}
\renewcommand{\arraystretch}{1}
\caption{ 
An overview of defective samples from the TILDA dataset. Red encirclement highlights the defects.}
\label{Fig.Tilda}
\end{figure}

Our methodology was additionally evaluated on the TILDA \cite{xie_improved_2021} textile datasets encompassing a diverse collection of 8 distinct textile types from  plain fabric to patterned fabric.
Various examples from defective samples are illustrated in Figure \ref{Fig.Tilda}. Results are depicted in Table \ref{Table 2}.

\begin{table*}[!h]
\captionsetup{font=small,justification=centering}
\caption{Anomaly detection results with AUROC on TILDA surfaces}
\resizebox{15.5cm}{!}{
\begin{tabular}{|c|ccc|c|} 
 \hline
		 \textbf{Category} & \textbf{PaDiM \cite{defard_padim_2020}}& \textbf{CFA \cite{lee_cfa_2022}} & 
        \textbf{Reverse distillation \cite{deng_anomaly_2022}} & \textbf{Ours} \\
        \hline
		{tilda1} &
		89.1 & 88.4  &  \textbf{94.8} & 90.2  \\
			
    	{tilda2} &
		  88.4 & 86.5  &  88.2 & \textbf{92.0}  \\
		
		{tilda3} &
		80.1 & 89.7  &  \textbf{91.4} & 84.8  \\
		
		{tilda4} &
		  45.9 & \textbf{83.6}  &  59.6 & 80.0  \\

            {tilda5} &
		  61.2 & 83.2  &  67.4 & \textbf{88.2}  \\

            {tilda6} &
		  79.1 & 85.7  &  78.7 & \textbf{93.0}  \\
    
            {tilda7} &
		  81.1 & \textbf{82.4}  &  78.6 & 79.7  \\
    
            {tilda8} &
		  45.8 & 48.1  &  \textbf{84.5} & 68.2  \\
		
		\hline
		{Mean} &
		71.3 & 80.9  &  80.4 & \textbf{84.5}  \\                                                    
		
	    \hline
\end{tabular}
}
\label{Table 2}
\end{table*}

\noindent The outcomes presented in Table \ref{Table 2} exemplify the competitiveness of our approach in comparison to other state-of-the-art methods. Our methodology showcases a mean Area Under the Receiver Operating Characteristic (AUROC) superior to alternative tested methods, and notably, it achieves a superior AUROC for 4 out of the 8 fabric types considered in the evaluation.

\subsection{Inference speed}

An essential advantage of our approach lies in its inference speed, which is primarily constrained by the selection of the pre-trained model employed for feature extraction. The architecture of the embedder, coupled with the straightforward comparison with one or a few clusters during inference, does not substantially increase the inference time. This critical advantage establishes our method as the fastest among counterparts employing the same pre-trained model. Furthermore, it stands out as a comparably swift solution even when compared to methods utilizing a smaller pre-trained model for feature extraction. This distinction is particularly noteworthy as such methods often incorporate a secondary model to extract pertinent anomaly detection information, thereby potentially introducing additional computational overhead.
An inference speed comparison is shown in Table \ref{TableSpeed}.

\begin{table*}[h]
	\centering
        \captionsetup{font=small,justification=centering}
	\footnotesize
	\setlength{\tabcolsep}{2pt}
        \caption{Comparison of pre-trained based approach in terms of inference time and frame per second}
        \resizebox{15.5cm}{!}{
	\begin{tabular}{|c|cccc|c|}
			
		\hline
		{\centering \textbf{Category}} & 
		{\textbf{PatchCore \cite{roth_towards_2021}}} &
		{\textbf{FastFlow \cite{yu_fastflow_2021}}} &
            {\textbf{RD\cite{deng_anomaly_2022}}} &
            {\textbf{RD \cite{deng_anomaly_2022}}} &
		{\textbf{Ours}} \\

            {\centering Extractor} & 
		{WideResnet50 \cite{zagoruyko_wide_2017} } &
		{WideResnet50 } &
            {WideResnet50 } &
            {Resnet18 \cite{he_deep_2015} } &
		{EfficientNet-b3 } \\
 
		\hline
		
		{FPS} &
		{\centering 5.8}  &
		{\centering 21.8} &
            {\centering 33}&
            {\centering 62}&
		{\centering 56}\\
		{Latency (ms)} &
		{\centering 172}  &
		{\centering 45.9} &
            {\centering 30}&
            {\centering 16}&
		{\centering 18}\\
		\hline
 		
	\end{tabular} 
        }

 \label{TableSpeed}
\end{table*}

\section{\uppercase{Ablation Study}}
\subsection{Comparison with a simple classifier}
To evaluate the effectiveness of our contrastive training approach, we conducted a comparative analysis with a traditional binary classifier. This classifier was trained on defect-free samples and artificially generated anomalous samples. We maintained consistency by extracting the same deep features from EfficientNet-B3. In contrast to our contrastive training methodology, the binary classifier was trained using standard binary classification techniques rather than adopting a contrastive learning framework. \\
The results obtained not only showcase the descriptive capability of the deep layers of EfficientNet but also affirm the superiority of our approach when compared to a straightforward classifier. 
It is noteworthy to highlight that the results achieved by the classifier remain highly impressive and are comparable to state-of-the-art methods from two years ago in the context of surface defect detection.
Results are shown in Table \ref{TableClassifier} for the surfaces of the MVTEC AD dataset.
\begin{table}[!h]
\captionsetup{font=small,justification=centering}
\caption{AUROC obtained a simple classifier trained on efficientNet-b3 deep features on MVTEC surfaces}
\resizebox{7.5cm}{!}{
\begin{tabular}{|c|ccccc|c|} 
 \hline
		 \textbf{category} & \textbf{carpet} & \textbf{wood}& \textbf{tile}&
        \textbf{leather} & \textbf{grid} & \textbf{mean}\\
        \hline
        {classifier} &
		99.2  & 99.1 & 98.0 & 100 & 94.5 &  98.2  \\
	    \hline
\end{tabular}
}
\label{TableClassifier}
\end{table}

\subsection{Decoder initialization and training}

As outlined in Section 3, we employ a decoder with frozen weights initialized randomly during the training process. While unconventional, we present our results with varying decoder initialization approaches: a decoder trained prior to embedder training, a decoder trained concurrently with the embedder, and a frozen decoder with random weights. Additionally, we offer an explanation for this unconventional methodology.
The results of the first aforementioned approach are presented in Table \ref{TableDecoder}.

\begin{table}[!h]
\captionsetup{font=small,justification=centering}
\caption{Anomaly detection results with AUROC on MVTEC surfaces}
\resizebox{7.5cm}{!}{
\begin{tabular}{|c|cccc|} 
 \hline
		 \textbf{Category} & \textbf{No decoder} & \textbf{Trained before} & \textbf{Trained together}& \textbf{Random} \\
        \hline
		{carpet} &
		  99.5 & 99.7  & 99.6 & 100  \\
			
    	{tile} &
		98.4 & 98.4  & 98.7 & 99.3  \\
		
		{wood} &
		99.9 & 100  & 99.9 & 100  \\
		
		{leather} &
		  100 & 100  & 99.9 & 100  \\

            {grid} &
		  99.3 & 99.6  & 98.4 & 99.6  \\
		
		\hline
		{Mean} &
		99.4 & 99.5  & 99.3 & 99.8  \\
		
	    \hline
\end{tabular}
}
\label{TableDecoder}
\end{table}

Our conjecture posits that confining the decoder's training exclusively to defect-free samples could induce a bias towards features crucial for reconstruction, potentially overlooking those essential for defect detection. This phenomenon results in a form of "concurrent" training between the embedder and the decoder. On the other hand, the random weight initialization provides a reconstruction with a statistically balanced mix of both representative features and those pertinent to defect detection. This randomness in reconstruction aligns optimally with our training objective.
An alternative option could have involved training the decoder on a combination of generated defective samples and defect-free samples. However, this approach yielded unsatisfactory results due to the limited training capacity of the decoder and the imperative for a compact architecture to ensure expeditious inference.

\subsection{Relevance of deep features}
\begin{figure}[!h]
\captionsetup{font=small,justification=centering}
\centerline{\includegraphics[scale=0.22]{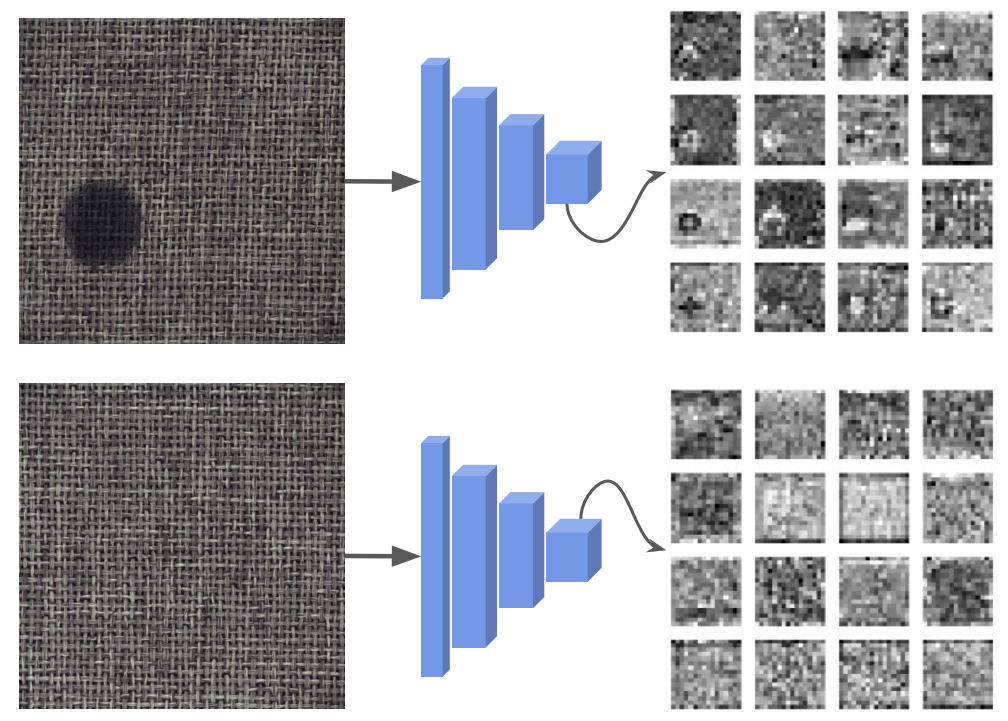}}
\renewcommand{\arraystretch}{1}
\caption{ 
A sample of features extracted from the layer of size 136x14x14 from EfficientNet-b3 }
\label{Fig5.1}
\end{figure}

In contrast to prevailing methodologies that utilize shallow and mid-level features from pre-trained models to mitigate bias towards specific classification tasks, our approach relies on deep features. These deep features, characterized by a lower resolution and a considerable number of filters, exhibit a pronounced bias toward classification making them unusable for object defect defection. 
This unconventional choice is elucidated by various considerations, encompassing the utilization of the contrastive loss function and the inherent characteristics of surface defect detection. 
In the context of a surface, a defect typically affects only a small portion while leaving the remainder unaffected. To optimize the effectiveness of the contrastive loss, it is advantageous to extract deep features where the defect, if discernible, occupies a more substantial portion of the feature space. This is achieved by employing deep features with a larger receptive field and lower resolution. Given that the defect constitutes a significant portion of the image, the contrastive loss methodology becomes particularly beneficial.
In contrast to objects, surfaces exhibit regularity, and the bias towards classification does not introduce misleading information. 
Indeed, as illustrated in Figure \ref{Fig5.1}, the features extracted from surfaces primarily capture regular patterns. However, when a defect emerges, it becomes readily discernible. 
These two considerations have been instrumental in guiding our decision regarding the selection of features.

\section{\uppercase{Conclusion}}
In this article, we introduced a novel method for unsupervised surface anomaly detection, centered around a contrastively selected embedding designed to aggregate the most pertinent features for the task of defect detection. Leveraging the representational capabilities of deep features extracted from a pre-trained model, our approach achieves state-of-the-art performance in surface defect detection on both the MVTEC AD dataset and the TILDA dataset. Through the employment of a compact network comprised of pointwise convolutions and a judicious selection of samples for inference comparison, our method ensures that inference speed is solely contingent on the chosen pre-trained classifier for deep feature extraction. This design leads to state-of-the-art performance in terms of model latency.
However, it is crucial to acknowledge the potential limitations of our method. The primary constraint is associated with the choice of the feature extractor and our substantial reliance on its representational power. As we focus on deep features, it becomes challenging to unbias the extracted features if the anomaly is not discernible within them. Another constraint lies in the process of defect generation during training, which significantly slows down model training, resulting in a relatively extended training duration compared to other state-of-the-art approaches.
In conclusion, we posit that this methodology holds considerable promise in the field of surface defect detection, and we earnestly encourage researchers to explore and further investigate such approaches.

\bibliographystyle{apalike}
{\small
\bibliography{Biblio}}

\end{document}